\documentclass[conference]{IEEEtran}
\IEEEoverridecommandlockouts

\usepackage{cite}
\usepackage{amsmath,amssymb,amsfonts}
\usepackage{algorithmic}
\usepackage{graphicx}
\usepackage{textcomp}
\usepackage{multirow}
\usepackage{subfig} 
\usepackage{booktabs} 
\usepackage{amsmath}  
\usepackage{xcolor}
\usepackage{xspace}
\usepackage{url}
\usepackage{hyperref}

\def\BibTeX{{\rm B\kern-.05em{\sc i\kern-.025em b}\kern-.08em
    T\kern-.1667em\lower.7ex\hbox{E}\kern-.125emX}}

\newcommand{\system}{\textsc{Vulcan}\xspace}

\begin{document}

\title{\system: Vision-Language-Model Enhanced Multi-Agent Cooperative Navigation for Indoor Fire-Disaster Response}

\author{\IEEEauthorblockN{Shengding Liu}
\IEEEauthorblockA{\textit{Computer Science \& Engineering} \\
\textit{Michigan State University}\\
East Lansing, MI, USA}
\and


\IEEEauthorblockN{Qiben Yan}
\IEEEauthorblockA{\textit{Computer Science \& Engineering} \\
\textit{Michigan State University}\\
East Lansing, MI, USA}
}

\maketitle

\begin{abstract}
Indoor fire disasters pose severe challenges to autonomous search and rescue due to dense smoke, high temperatures, and dynamically evolving indoor environments. In such time-critical scenarios, multi-agent cooperative navigation is particularly useful, as it enables faster and broader exploration than single-agent approaches. However, existing multi-agent navigation systems are primarily vision-based and designed for benign indoor settings, leading to significant performance degradation under fire-driven dynamic conditions. In this paper, we present \system, a multi-agent cooperative navigation framework based on multi-modal perception and vision–language models (VLMs), tailored for indoor fire disaster response. We extend the Habitat-Matterport3D benchmark by simulating physically realistic fire scenarios, including smoke diffusion, thermal hazards, and sensor degradation. We evaluate representative multi-agent cooperative navigation baselines under both normal and fire-driven environments. Our results reveal critical failure modes of existing methods in fire scenarios and underscore the necessity of robust perception and hazard-aware planning for reliable multi-agent search and rescue.
\end{abstract}


\section{Introduction}
Natural and human-made disasters pose immediate threats to life and infrastructure, requiring rapid and coordinated emergency response. In hazardous environments such as indoor fires, high temperature, dense smoke, and structural instability make human intervention extremely dangerous. 
A recent devastating fire in Tai Po, Hong Kong~\cite{Grundy2025TaiPoFire} greatly impeded human access and situational awareness during the early response phase, leading to significant casualties and property damage. It was the deadliest fire in Hong Kong in decades.

Mobile robots have emerged as a promising alternative for operating in life-threatening environments. By enabling autonomous exploration, hazard perception, and target search under severe perception degradation, robotic systems can provide rapid situational awareness while reducing the risk to human responders and supporting more informed decision-making.

\begin{figure}
    \centering
    \includegraphics[width=1.0\linewidth]{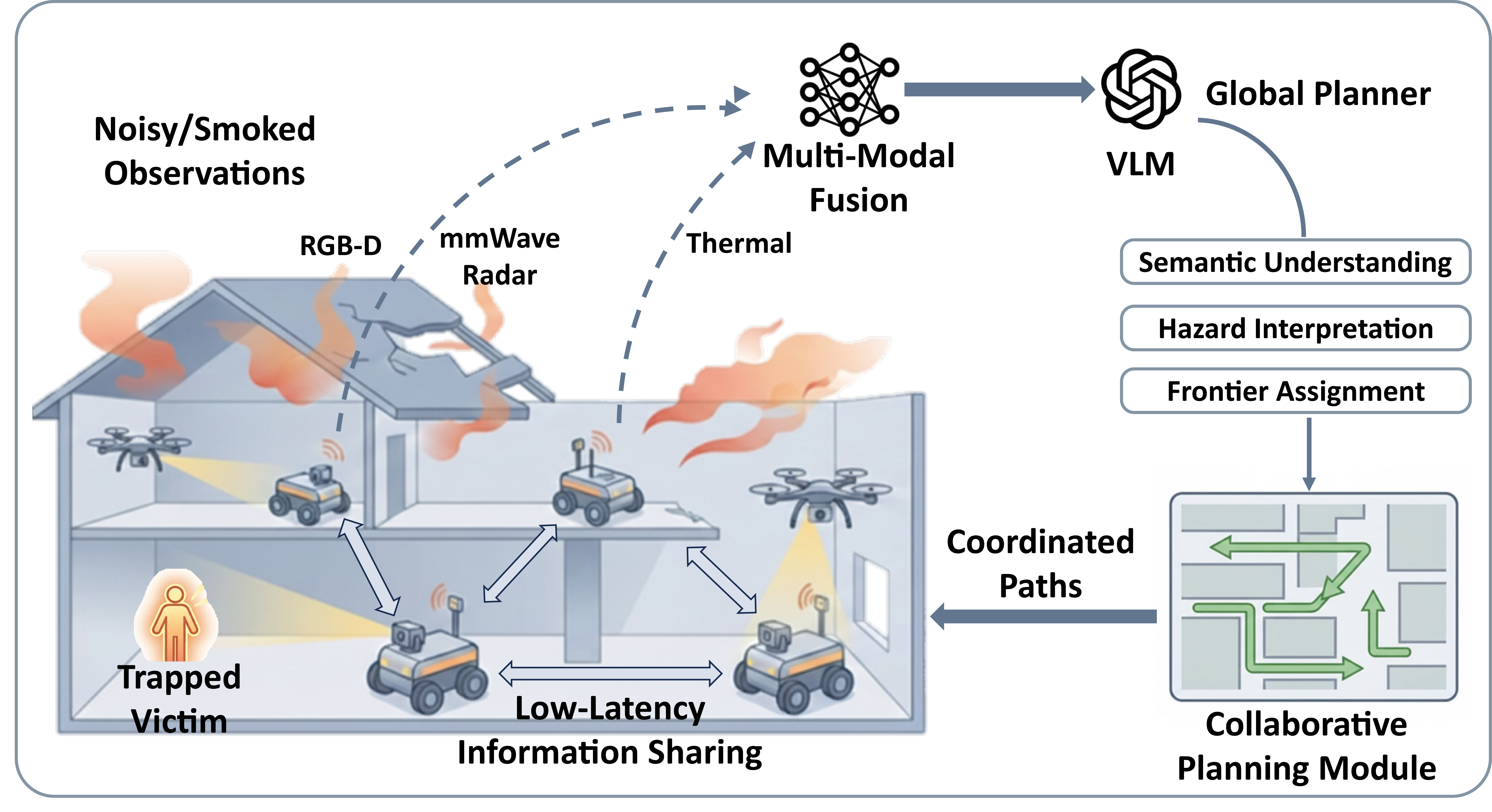}
    \caption{Overview of \system for multi-agent search and rescue in indoor fire scenarios.}
    \label{fig:scenario}
    \vspace{-15pt}
\end{figure}

Despite recent progress, most existing search and rescue (SAR) deployments~\cite{queralta2008collaborative} still rely on \emph{single-robot exploration} or loosely coordinated multi-robot routines, where individual robots operate with minimal information sharing. Such strategies inherently limit system-level efficiency, particularly in large, unmapped, or dynamically evolving indoor environments, thereby underscoring the urgent need for multi-agent cooperative navigation. 
Fig.~\ref{fig:scenario} shows an example scenario of multi-agent search and rescue in an indoor fire environment.
\begin{figure}[htbp]
    \centering

    \subfloat[RGB camera]{
        \includegraphics[width=0.45\linewidth]{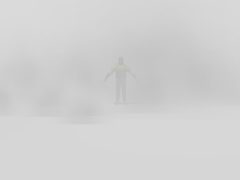}
        \label{fig:rgb}
    }
    \hfill
    \subfloat[Depth camera]{
        \includegraphics[width=0.45\linewidth]{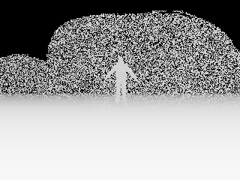}
        \label{fig:depth}
    }


    \subfloat[Lidar]{
        \includegraphics[width=0.45\linewidth]{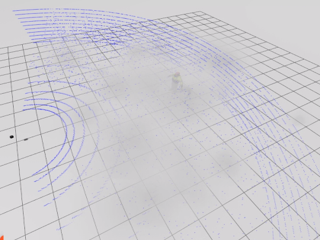}
        \label{fig:lidar}
    }
    \hfill
    \subfloat[Thermal camera]{
        \includegraphics[width=0.45\linewidth]{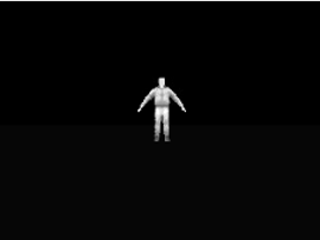}
        \label{fig:thermal}
    }

    \caption{Gazebo-based fire simulation demonstrates the degradation of modality-dependent perception  as smoke density increases.}
    \vspace{-15pt}
    \label{fig:four_sensors}
\end{figure}

In recent years, vision-based cooperative navigation has attracted increasing attention, driven by advances in simulation platforms, 
and large-scale 3D scene datasets~\cite{ramakrishnan2021habitat}. Existing approaches to visual navigation are typically grouped into two main paradigms: \emph{end-to-end} systems~\cite{khandelwal2022simple}, which directly map sensory observations to control actions, and \emph{modular pipelines}~\cite{ramakrishnan2022poni}, which explicitly decompose the task into perception, mapping, planning, and control. Both paradigms have demonstrated effectiveness by leveraging spatial and semantic representations learned via supervised or reinforcement learning. Furthermore, recent work has increasingly explored the integration of vision–language
models (VLMs) into visual navigation systems~\cite{11302789}.
By jointly reasoning over visual observations and natural language, VLMs introduce powerful semantic priors that enhance object grounding, relational reasoning, and holistic understanding of complex indoor environment. 

However, the majority of existing multi-agent cooperative navigation frameworks~\cite{11302789,liu2024air,yu2022learning} are predominantly vision-based and developed under assumptions of static, well-structured, and relatively benign indoor environments. Such assumptions fundamentally break down in fire-driven indoor scenarios, where perception is severely degraded by smoke and high temperatures, environmental conditions evolve rapidly and unpredictably, and inefficient or uncoordinated exploration can directly compromise both agent safety and mission success. These problems highlight a critical gap in current navigation frameworks and motivate the need for an efficient and safe cooperative multi-agent navigation tailored for indoor fire-disaster response.

\begin{figure*}[t]
  \centering
  \includegraphics[width=\textwidth]{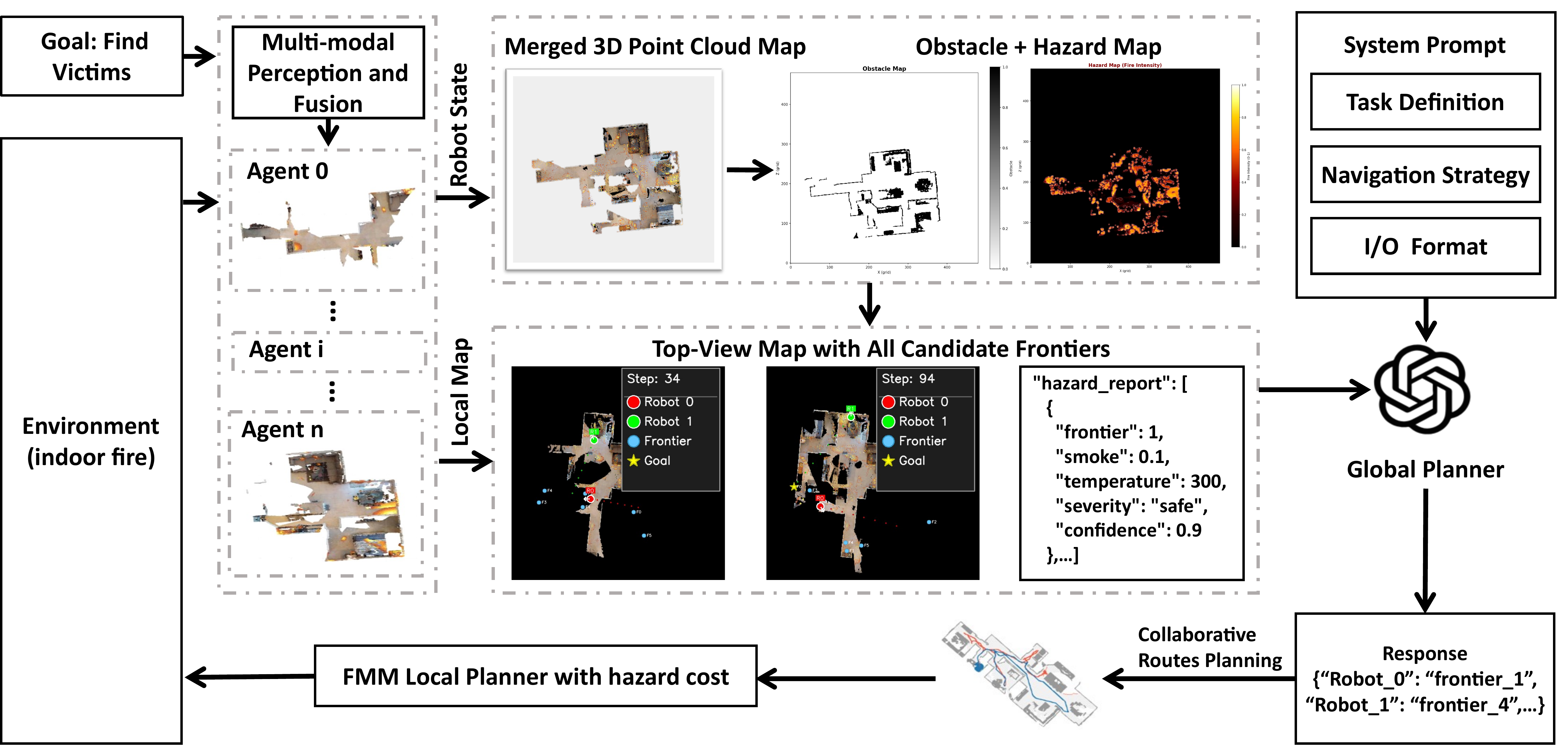}
  \caption{System architecture of \system. Each agent performs multi-modal perception and fusion, constructs hazard-aware global maps, and plans safe and efficient exploration using a VLM-based global planner and hazard-aware FMM local planner.}
  \label{fig:System Design}
  \vspace{-15pt}
\end{figure*}

Despite recent advancements in cooperative navigation including learning-based approaches~\cite{khandelwal2022simple}, VLM powered planning~\cite{11302789}, and embodied multi-agent coordination frameworks~\cite{tan2025roboos}, directly applying these approaches to fire-driven indoor scenarios remains extremely challenging. Compared to static or benign environments, indoor fire scenes introduce several fundamental challenges:

\subsubsection{Dynamic and rapidly evolving environments} 
Temperature gradients, flame propagation, and smoke diffusion continuously reshape the environment, altering traversability, visibility, and semantic cues over time.

\subsubsection{Unreliable and degraded sensor data} 
Fire-induced smoke severely degrades multi-modal perception in indoor environments. 
As illustrated in Fig.~\ref{fig:four_sensors}, our Gazebo-based fire simulations, built on a high-fidelity physics engine with particle-emitter support, qualitatively demonstrate modality-dependent perception degradation under increasing smoke density. 

\subsubsection{Latency-critical and hazard aware decision making} 
SAR missions demand real-time coordination, as excessive planning delays or inefficient multi-robot routing can directly compromise human survival and overall mission success. Simultaneously, it is crucial to ensure that each agent can navigate and explore safely, avoiding hazards while completing its tasks.

To address these challenges, we introduce \textbf{\system}\footnote{\system is the Roman deity of fire.}, a hazard-aware multi-agent cooperative navigation framework tailored for indoor fire SAR scenarios. As illustrated in Fig. ~\ref{fig:scenario}, each agent performs multi-modal perception and fusion to jointly construct a hazard-aware global map. Based on this shared representation, exploration is coordinated through a VLM-based global planner and a hazard-aware local planner to ensure safe and efficient navigation. While a complete end-to-end deployment is ongoing, we implement key components and present preliminary evaluations demonstrating the framework’s feasibility and the limitations of conventional multi-agent navigation in fire scenarios. Live demonstrations and representative results are available at \url{https://vulcan-dr.github.io}.


To summarize, our contributions are as follows:
\begin{itemize}
    \item We propose \system, a hazard-aware multi-agent cooperative navigation framework tailored for indoor fire scenarios, which maintains high navigation efficiency despite severe fire-induced perception degradation and environmental hazards.
    \item We extend indoor navigation benchmarks with realistic fire simulations and coordinate a VLM-based global planner with a hazard-aware local controller.
    \item We evaluate representative multi-agent navigation baselines under both normal and simulated indoor fire conditions, and systematically analyze the underlying causes of their performance degradation in fire scenarios.
\end{itemize}

\section{System Design}
\subsection{Task Definition}
We consider a \emph{multi-agent cooperative SAR} task in fire-driven indoor environments, where a team of mobile robots explores an unknown scene to locate victims or mission-critical targets under hazardous conditions. Let $\mathcal{C} = \{c_1, \ldots, c_m\}$ denote the set of target categories (e.g., \emph{human}, \emph{fire extinguisher}, \emph{bathroom}), and let $\mathcal{S} = \{s_1, \ldots, s_K\}$ denote the set of fire-affected indoor scenes. In each episode, $N$ robots $\mathcal{R} = \{r_1, \ldots, r_N\}$ are deployed in a scene $s_i \in \mathcal{S}$ from a shared starting position $\mathbf{p}_i$ with different initial orientations to reflect realistic emergency deployment. All robots are assigned the same target category $c_i \in \mathcal{C}$. The episode is defined as:
\begin{equation}
\mathcal{T}_i = \{N, s_i, c_i, \mathbf{p}_i\}.
\end{equation}

The observation space $\mathcal{O}$ comprises four modalities: RGB, depth, thermal imaging, and mmWave radar, providing complementary visual, geometric, and smoke-robust perception. The action space $\mathcal{A}$ consists of six discrete actions:
\{\texttt{move\_forward}, \texttt{turn\_left}, \texttt{turn\_right}, \texttt{look\_up}, \texttt{look\_down}, \texttt{stop}\}. The \texttt{move\_forward} action advances the robot by $0.25\,\text{m}$, and each rotational action changes orientation by $30^{\circ}$. At each discrete time step $t$, robot $r_i$ receives a multi-modal observation $o_t^i \in \mathcal{O}$ and executes an action $a_t^i \in \mathcal{A}$. An episode is successful if the team reaches the required number of target instances of category $c_i$ within a $0.1\,\text{m}$ distance threshold, or fully explores the environment, while all agents remain within survivable hazard limits throughout the episode.

\subsection{Overview}
As illustrated in Fig.~\ref{fig:System Design}, we propose a hazard-aware multi-agent navigation framework, \system, tailored for indoor fire SAR scenarios. Each agent performs robust multi-modal perception by jointly fusing RGB-D, thermal imaging, and mmWave radar to see through smoke and construct a local hazard-annotated 3D point cloud encoding geometry, temperature, smoke and sensing uncertainty. These local maps are merged into a global representation, which is further projected into a compact 2D obstacle and hazard map for frontier extraction and risk assessment. Leveraging this structured spatial and hazard-aware abstraction, a VLM serves as a global planner, reasoning jointly over robot states, frontier utility, and environmental risks to assign safe and efficient exploration goals to each agent. Finally, a hazard-aware Fast Marching Method (FMM) local planner refines these assignments into collision-free and risk-sensitive trajectories. Together, this hierarchical design enables coordinated, safe, and efficient multi-agent exploration under dynamic, fire-driven indoor conditions.

\subsection{Multi-modal Perception and Fusion}
\begin{figure}[t]
  \centering
  \includegraphics[width=\columnwidth]{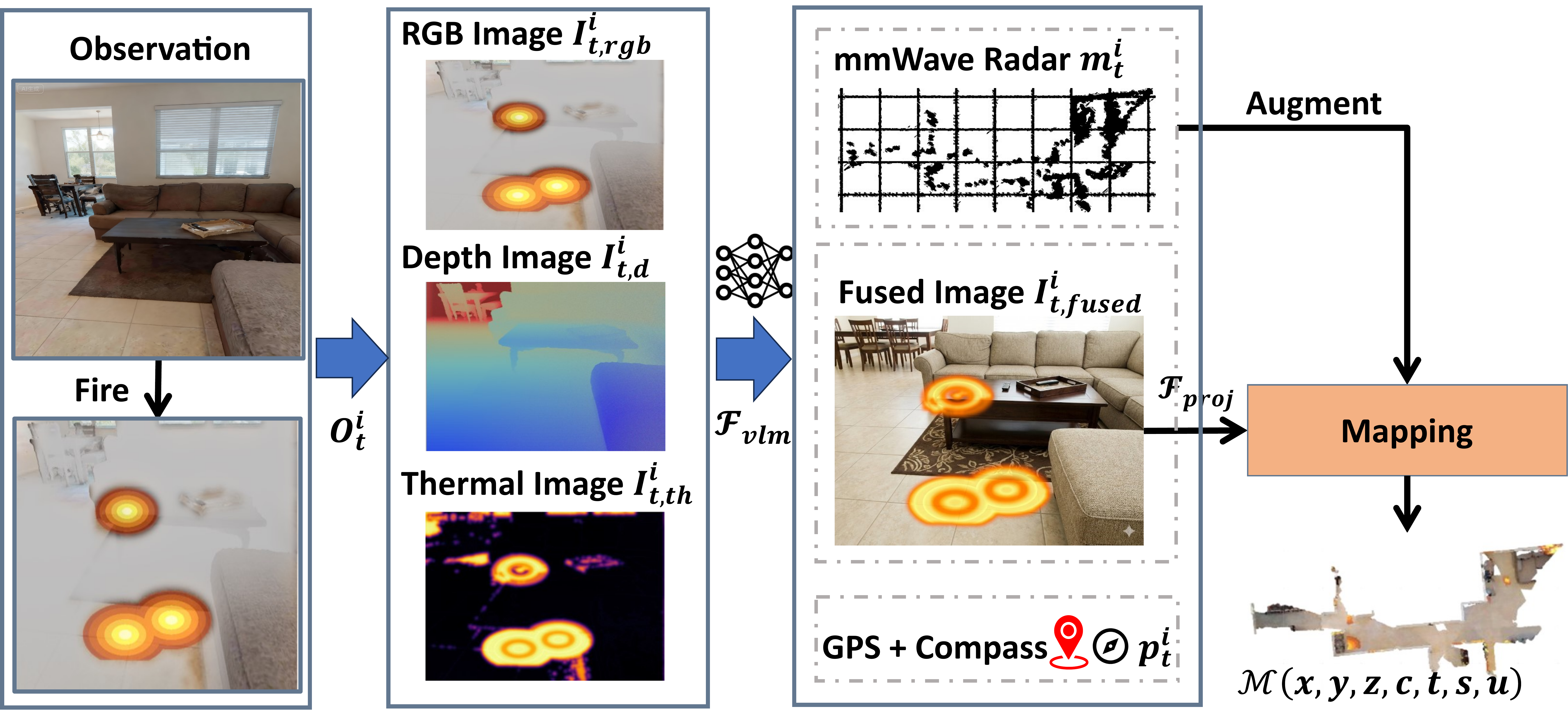}
  \caption{Architecture of multi-modal perception and cross-modal fusion for robust environment understanding.}
  \label{fig:multi_modal_perception_and_fusion}
  \vspace{-15pt}
\end{figure}

To maintain reliable situational awareness under such harsh conditions, our framework adopts a robust multi-modal perception pipeline that integrates RGBD, thermal imaging, and mmWave radar signals as illustrated in Fig.~\ref{fig:multi_modal_perception_and_fusion}. The goal is to (1) extract stable structural cues, (2) detect fire-related hazards, and (3) produce a unified hazard-aware map for downstream planning.

For each robot $r_i$, given a sequence of synchronized multi-modal observations collected in indoor fire environments,
including RGB, depth, and thermal images,
\begin{equation}
\mathcal{I}_i =
\left\{
\left(I^{i}_{1,\mathrm{rgb}}, I^{i}_{1,\mathrm{d}}, I^{i}_{1,\mathrm{th}}\right), \dots,
\left(I^{i}_{t,\mathrm{rgb}}, I^{i}_{t,\mathrm{d}}, I^{i}_{t,\mathrm{th}}\right)
\right\},
\end{equation}
and the corresponding mmWave radar observations and the associated robot poses are:
\begin{equation}
\mathcal{M}_i^{\mathrm{rad}} =
\left\{
m_1^i, \dots, m_t^i
\right\}, \quad
\mathcal{P}_i = \{p^i_1, \dots, p^i_t\}.
\end{equation}
Each robot incrementally constructs a local hazard-aware 3D point cloud map $\mathcal{M}_i$ through the multi-modal perception and its pose.

Specifically, at each time step $t$, three visual modalities are first spatially and temporally aligned.
The aligned observations are then jointly processed by a VLM-based fusion operator $\mathcal{F}_{\mathrm{vlm}}(\cdot)$ to infer an enhanced visual representation:
\begin{equation}
I^{i}_{t,\mathrm{fused}} =
\mathcal{F}_{\mathrm{vlm}}
\left(
I^{i}_{t,\mathrm{rgb}},
I^{i}_{t,\mathrm{d}},
I^{i}_{t,\mathrm{th}}
\right).
\end{equation}
The resulting fused image $I^{i}_{t,\mathrm{fused}}$ approximates a \emph{smoke-transparent} view of the environment, enabling more reliable downstream tasks such as 3D reconstruction, hazard mapping, and frontier extraction.

An open-vocabulary detector $\mathrm{Det}(\cdot)$ is then applied to the fused image $I^i_{t,\mathrm{fused}}$ to identify candidate objects and fire-related regions. A class-agnostic segmentation model $\mathrm{Seg}(\cdot)$ further extracts pixel-wise masks for these regions. Each pixel $u$ in the fused image, together with its depth $d(u)$, is back-projected to a 3D point in the camera frame: 
\begin{equation}
\mathbf{x}_c(u) = d(u)\mathbf{K}^{-1}\tilde{u},
\end{equation}
where $\mathbf{K}$ is the camera intrinsic matrix and $\tilde{u}$ is the homogeneous pixel coordinate.

Thermal measurements are aligned with RGB pixels to estimate temperature $T(u)$, while smoke density $S(u)$ is inferred from RGB degradation and depth consistency. Each point is augmented with a hazard attribute vector:
\begin{equation}
\mathbf{h} = [T(u), S(u), \sigma(u)],
\end{equation}
where $\sigma(u)$ encodes sensing uncertainty due to occlusion, smoke, or missing depth returns.

The local 3D hazard-aware point cloud is then formed by combining the vision-based points with mmWave radar points:
\begin{equation}
\mathcal{M}_t^{i} =
\mathcal{F}_{\mathrm{proj}} \Big( 
    \mathrm{Seg} \big( \mathrm{Det}(I^i_{t,\mathrm{fused}}) \big) 
\Big) 
\;\cup\; 
m_t^i.
\end{equation}

The global hazard-aware point cloud map $\mathcal{M}$ is then constructed by jointly aggregating vision-based hazard points and mmWave radar points from all robots into a common coordinate frame:
\begin{equation}
\mathcal{M} =
\bigcup_{i=1}^{n}
\;\bigcup_{\tau=1}^{t}
\mathcal{M}_\tau^{i}
\label{eq:global_map}
\end{equation}

To suppress noisy measurements introduced by fire flickering, smoke diffusion, and thermal sensor artifacts, we apply DBSCAN clustering to the merged point cloud and remove sparse outliers.

\subsection{Global Map Construction and Frontier Risk Assessment}
\label{section: Global Map Construction and Frontier Risk Assessment}

To enable efficient risk-aware exploration in fire scenarios, a global 2D obstacle–hazard map is constructed from the 3D point cloud observations collected by the robots. The map provides a compact and physically grounded representation of both structural constraints and environmental hazards.

\subsubsection{2D Obstacle-Hazard Map Construction}

At the beginning of each episode, the global 3D map $\mathcal{M}$ is projected into a 2D grid-based representation centered at the robot's initial position. The resulting map consists of two aligned channels: an obstacle map $\mathbf{O}$ and a hazard map $\mathbf{H}$.

The obstacle map is constructed via top-down projection of non-floor 3D points, where grid cells exceeding a point-density threshold are marked as occupied. This produces a binary occupancy layout used to constrain navigation. The hazard map encodes fire-related environmental risks by aggregating hazard attributes from projected 3D points. Specifically, the hazard intensity of each grid cell is computed as a weighted combination of temperature, smoke density, and sensing uncertainty:
\begin{equation}
\mathbf{H}(x, y) = \sum_{x_g \in (x, y)} \left( w_T T + w_S S + w_\sigma \sigma \right),
\end{equation}
where $w_T$, $w_S$, and $w_\sigma$ balance the contributions of different hazard factors. Together, $\mathbf{O}$ and $\mathbf{H}$ form a compact obstacle–risk abstraction of the fire scene.

\subsubsection{Frontier Extraction and Risk Assessment}

Frontiers are defined as the boundaries between explored and unexplored regions after removing obstacle boundaries from $\mathbf{O}$. Frontier cells are clustered via connected-component analysis, and small noisy clusters are discarded.

Each frontier is assigned a risk score by aggregating hazard intensities within its region. Let $\mathcal{F}_k$ denote the set of grid cells belonging to frontier $k$. The frontier-level risk is defined as:
\begin{equation}
R_k = \frac{1}{|\mathcal{F}_k|} \sum_{(x,y) \in \mathcal{F}_k} \mathbf{H}(x,y).
\end{equation}

Frontiers are then categorized into qualitative risk levels (e.g., safe, moderate, dangerous) using predefined thresholds. Frontiers with lower risk scores are prioritized by the VLM-based planner to enable safe and efficient exploration in fire-disaster environments.

\subsection{VLM-Based Global Planning under Fire Hazards}
After the global map construction and frontier risk assessment, we employ a VLM as a global planner to assign the frontiers to each agent through structured multi-modal prompts.

\subsubsection{Multi-Modal Prompt Design}
The VLM receives a compact yet expressive prompt consisting of a visual input and a textual input.

\emph{Visual prompt.}
The visual input is a top-down global map rendered from the merged hazard-aware 3D point cloud and aligned with the 2D obstacle--hazard exploration map. Robot positions and identifiers are explicitly marked, while all candidate frontiers are highlighted and labeled. This representation integrates geometric structure, semantic context, and multi-robot configuration for global spatial reasoning. An example of the visual prompt is shown in Fig.~\ref{fig:System Design}.

\emph{Textual prompt.}
The textual input consists of a system prompt defining the fire-response navigation task and a frontier-level hazard report in JSON format. The system prompt specifies the mission objective, multi-robot setting, map semantics, and output constraints, while the hazard report summarizes estimated smoke density, temperature, hazard severity, and confidence for each frontier.

\subsubsection{Global Frontier Assignment}
Following prior work \cite{11302789}, we restrict the VLM to high-level reasoning and decouple it from low-level motion control. 
At each global planning step, the VLM receives the updated multi-modal prompt and assigns a long-term frontier goal to each robot. The assignment balances two objectives: (i) maximizing exploration efficiency by reducing inter-robot redundancy, and (ii) minimizing fire exposure by favoring lower-risk frontiers with reliable perception. If no safe frontier exists, fallback goals are selected within explored low-risk regions.

\begin{table}[t]
    \centering
    \caption{Performance comparison under normal conditions}
    \label{tab:normal_condition}
    \begin{tabular}{l c c c c}
        \toprule
        Method & NS & SR & SPL  & CHE\\
        \midrule
        Greedy~\cite{visser2013discussion}       
        & 219.03 & 0.686 & 0.322 & 0 \\
        Cost-Utility~\cite{julia2012comparison}   
        & 199.60 & 0.628 & 0.315 & 0\\
        Random Sample   
        & 206.62 & 0.631 & 0.258 & 0 \\
        Co-NavGPT~\cite{11302789}    
        & 185.43 & 0.666 & 0.388 & 0 \\
        \bottomrule
    \end{tabular}
\end{table}

\begin{table}[t]
    \centering
    \caption{Performance comparison under fire conditions}
    \label{tab:fire_condition}
    \begin{tabular}{l c c c c c}
        \toprule
        Method & NS$\uparrow$ & SR$\downarrow$ & SPL$\downarrow$ & CHE$\uparrow$ \\
        \midrule
        Greedy        
        & 267.40 & 0.651 & 0.319 & 11.233 \\
        Cost-Utility  
        & 207.49 & 0.608 & 0.306 & 8.517 \\
        Random Sample 
        & 214.96 & 0.625 & 0.251 & 7.174 \\
        Co-NavGPT   
        & 187.89 & 0.660 & 0.381 & 4.873 \\
        \bottomrule
    \end{tabular}
    \vspace{-15pt}
\end{table}

\begin{figure*}[t]
    \centering

    \subfloat[Missed detection before (left figure) and after (right figure) fire scenario.]{
        \includegraphics[width=0.382\textwidth]{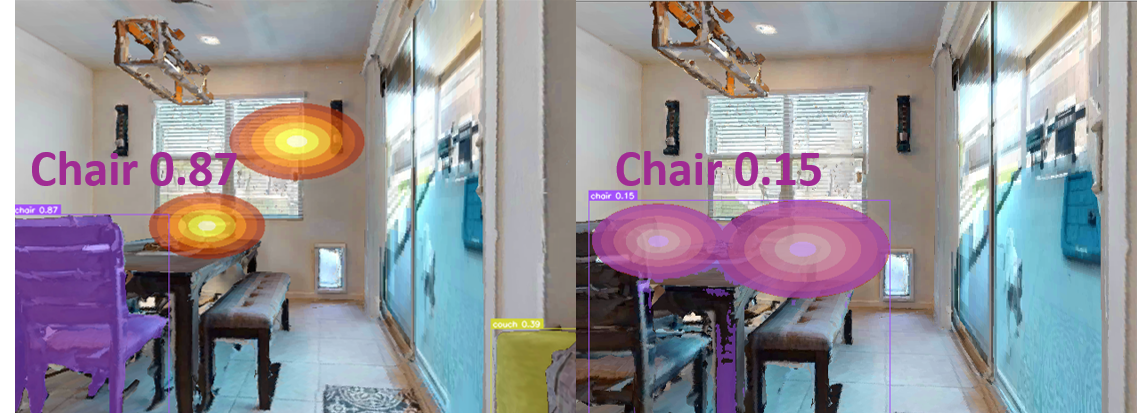}
    }
    \hfill
    \subfloat[Exploration in normal scenario.]{
        \includegraphics[width=0.29\textwidth]{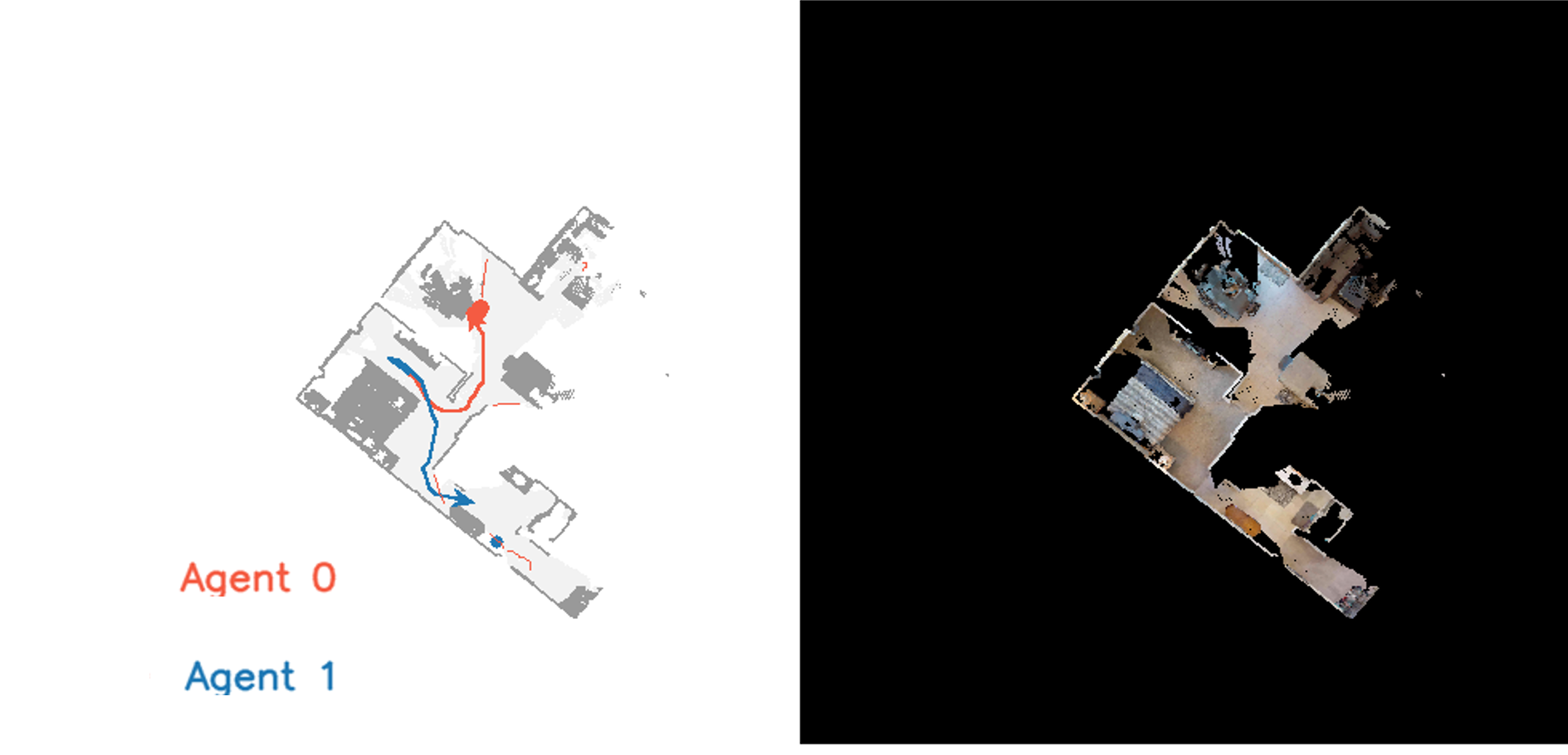}
    }
    \hfill
    \subfloat[Hazard map.]{
        \includegraphics[width=0.283\textwidth]{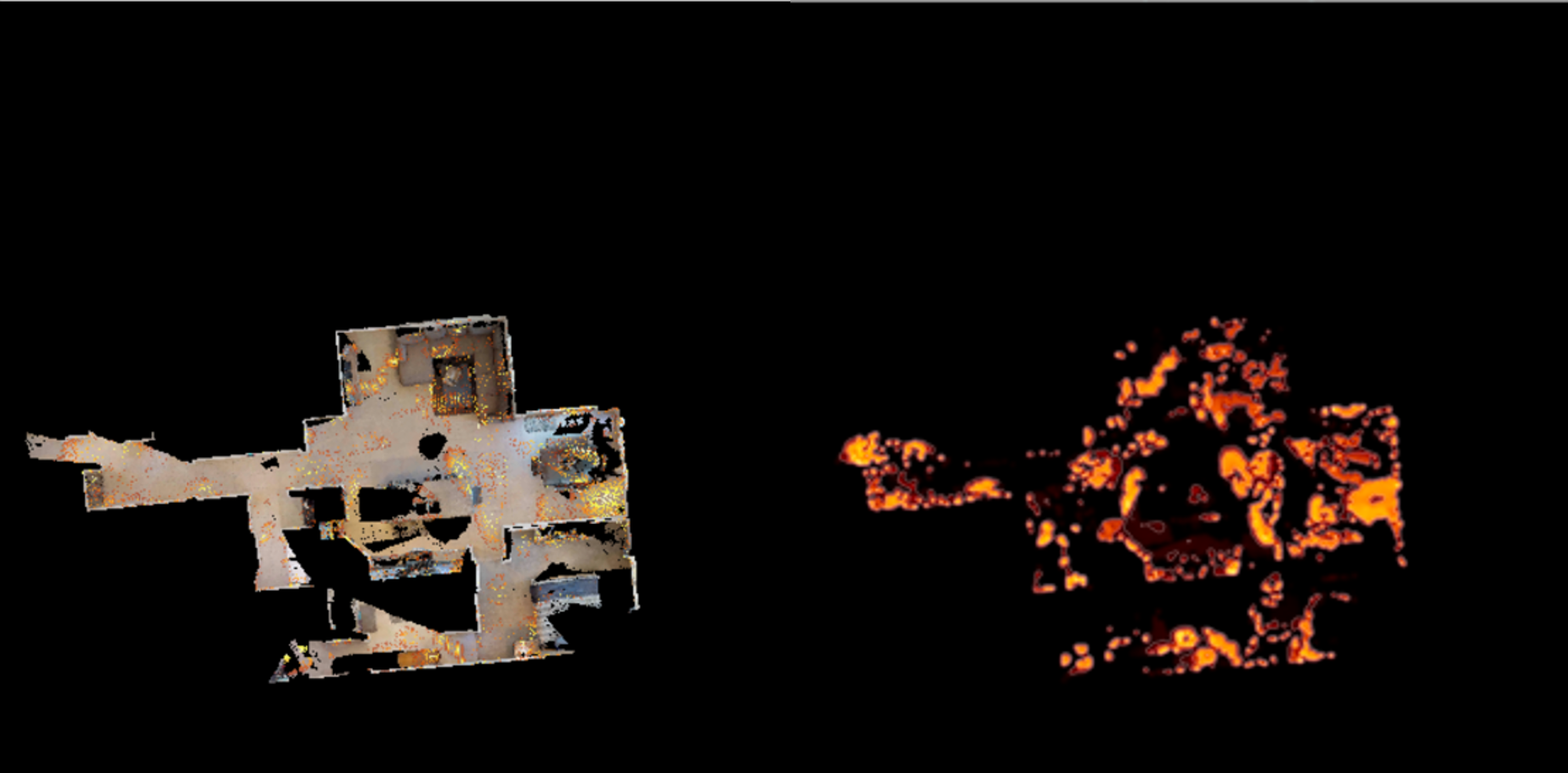}
    }


    \subfloat[False detection before (left figure) and after (right figure) fire scenario.]{
        \includegraphics[width=0.382\textwidth]{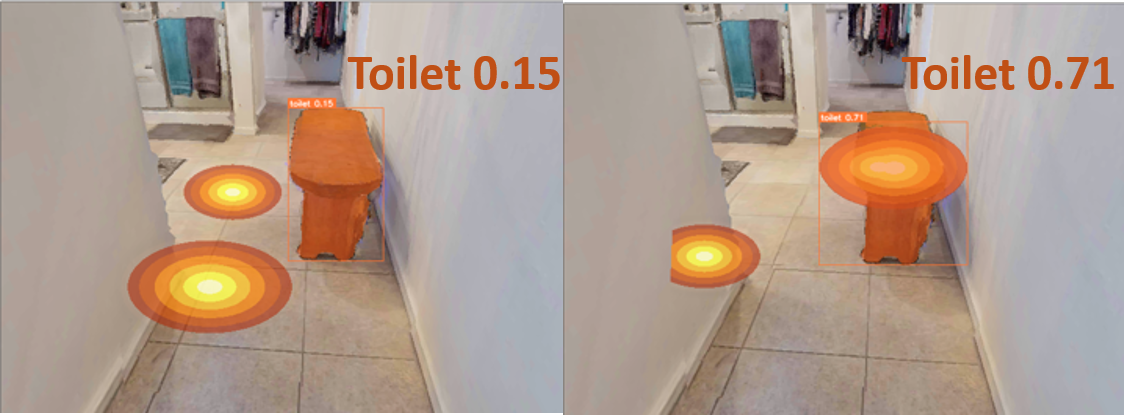}
    }
    \hfill
    \subfloat[Exploration in fire scenario.]{
        \includegraphics[width=0.29\textwidth]{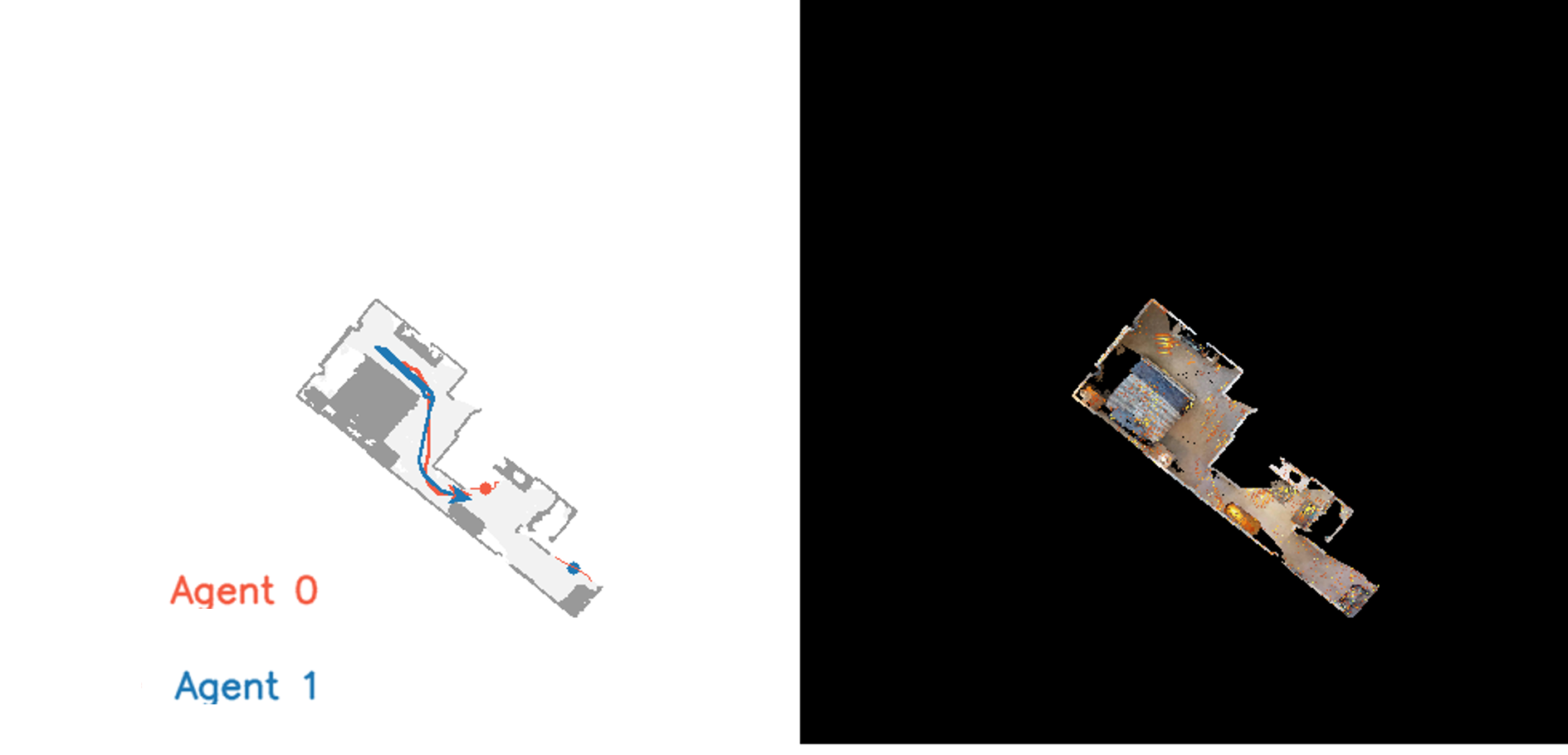}
    }
    \hfill
    \subfloat[Dangerous routing scenario.]{
        \includegraphics[width=0.283\textwidth]{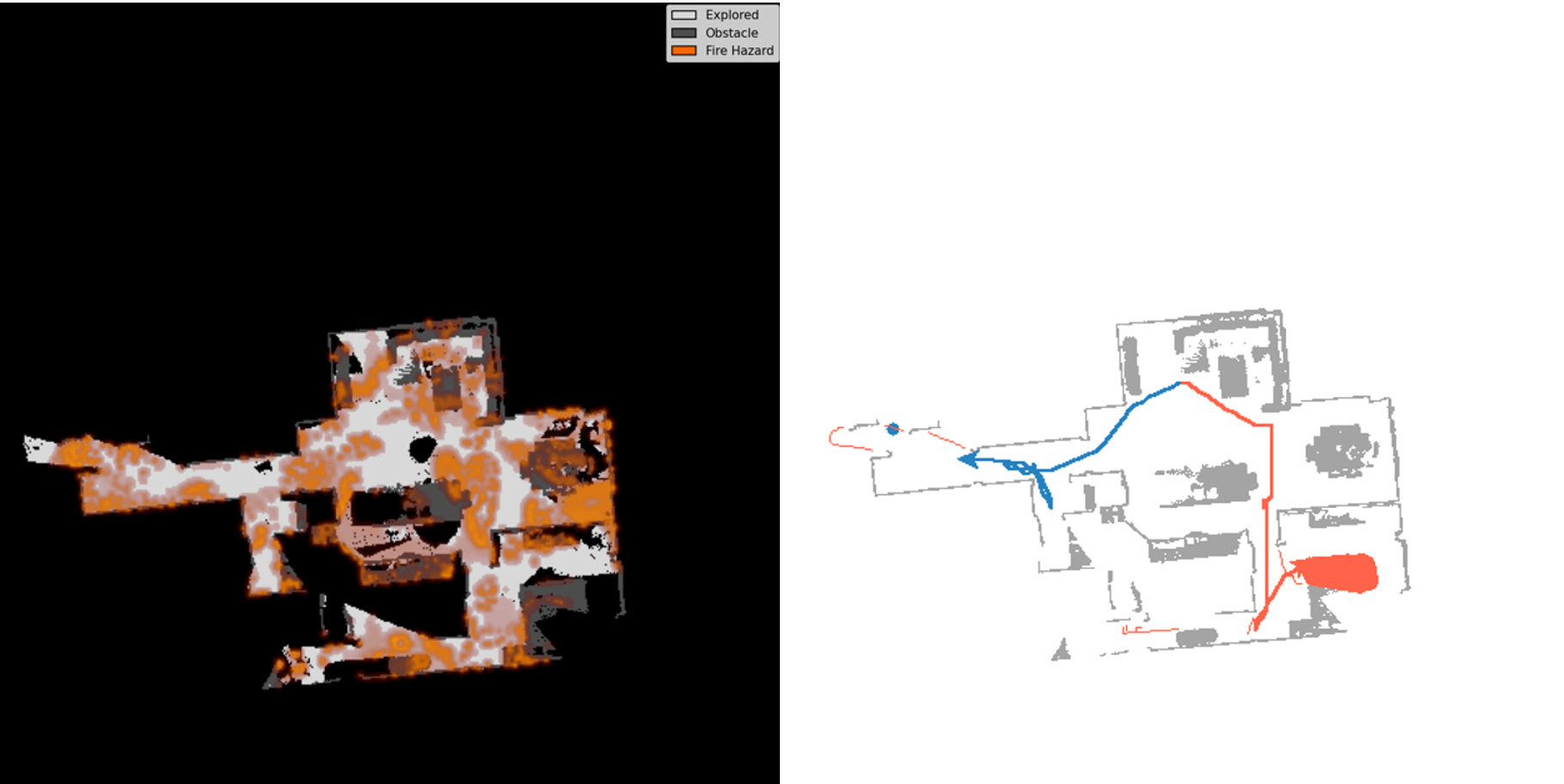}
    }

    \caption{The comparison highlights three failur1e modes: (a, d) Perception failure: smoke causes missed detections (confidence drop in a) and false positives (hallucinations in d); (b, e) Inefficient exploration: agents exhibit redundant framing and reduced coverage under fire conditions; (c, f) Unsafe planning: without hazard awareness, agents plan paths through high-risk zones (f) despite the presence of thermal hazards shown in the ground-truth map (c).}
    \label{fig:evaluate_fire}
    \vspace{-15pt}
\end{figure*}

\subsection{Hazard-Aware Fast Marching Method for Local Planning}

Given the global 2D obstacle and hazard maps, local navigation in fire-driven indoor environments must jointly consider geometric feasibility and environmental risk. We adopt a hazard-aware FMM as the local planner, enabling real-time generation of smooth, collision-free, and risk-sensitive trajectories.

To incorporate fire risk, we modulate the FMM propagation speed using the hazard map:
\begin{equation}
    F(x) = \frac{1}{1 + \alpha H(x)},
\end{equation}
where $H(x)$ encodes aggregated fire-related risk, and $\alpha$ controls the trade-off between safety and efficiency. This formulation penalizes traversal through high-risk regions even when they are geometrically free, while allowing fast propagation in safer areas. 

Applying FMM with the hazard-modulated speed produces an arrival-time field that naturally biases paths toward low-risk corridors. The local trajectory is extracted by following the negative gradient of this field, yielding a smooth path that minimizes accumulated risk while respecting obstacle constraints.


\section{Evaluation}
In this section, we evaluate representative multi-agent, map-based navigation baselines, including both VLM-based planners and conventional map-based methods under normal conditions and simulated indoor fire conditions. 
\subsection{Datasets}
To enable a fair and realistic comparison, we extend the Habitat-Matterport3D~\cite{ramakrishnan2021habitat} dataset by injecting fire-specific environmental degradations, including varying levels of smoke density, flame intensity, and heat-induced visibility loss. These modifications emulate the perceptual and navigational challenges commonly encountered in real-world indoor fire scenarios. Following the standard protocol, we use 36 validation scenes containing 200 episodes with semantic object annotations. We select six goal categories: chair, sofa, plant, bed, toilet, and TV. Since the dataset contains no human instances, we use these objects as proxies to simulate human presence.
\subsection{Metrics}
 Our evaluation considers not only task completion but also safety related constraints imposed by fire hazards.
We evaluate all the methods using the following four metrics: Number of Steps (NS), Success Rate (SR), Success weighted by Path Length (SPL), and Cumulative Hazard Exposure (CHE). NS measures exploration efficiency and is defined as the total number of discrete time steps taken before episode termination. The SR is defined as:
\begin{equation}
    \mathrm{SR} = \frac{1}{N} \sum_{i=1}^{N} S_i ,
\end{equation}
and SPL is defined as:
\begin{equation}
    \mathrm{SPL} = \frac{1}{N} \sum_{i=1}^{N} 
    S_i \frac{l_i}{\max(l_i, p_i)} ,
\end{equation}
where $N$ is the number of episodes, $S_i \in \{0,1\}$ indicates success in episode $i$, $l_i$ denotes the shortest path length from the start to any success location, and $p_i$ is the shortest trajectory length executed by any robot in episode $i$. To quantify navigation safety, CHE measures the cumulative risk experienced by all robots and is defined as: 
\begin{equation}
    \mathrm{CHE} = \sum_{i=1}^{N} \sum_{a \in \mathcal{A}} \sum_{t=1}^{T_i}
    H\bigl(x_{a,t}^{(i)}\bigr),
\end{equation}
where $\mathcal{A}$ denotes the robot set, $T_i$ is the episode length, $x_{a,t}^{(i)}$ is the position of robot $a$ at time step $t$, and $H(\cdot)$ is the hazard intensity from the global obstacle--hazard map. Lower CHE indicates safer exploration with reduced exposure to fire-related hazards.
\subsection{Results and Analysis}

We evaluate all navigation methods under both normal indoor conditions and fire-driven scenarios to assess their navigation success rate, exploration efficiency, and safety awareness. The results are summarized in Table~\ref{tab:normal_condition} and Table~\ref{tab:fire_condition}. 

Comparisons among different methods within the same environment show that 
the VLM-based planner (Co-NavGPT) achieves higher success rates and SPL while requiring fewer steps and significantly reducing cumulative hazard exposure. These results indicate that high-level semantic reasoning enables agents to assign spatially efficient and semantically meaningful exploration goals, thereby improving coordination and reducing redundant exploration.

Furthermore, by comparing the performance of each method across normal and fire scenarios, we observe a substantial performance degradation for existing approaches originally designed for clean indoor environments. 
We analyze the underlying causes of the observed performance degradation under fire-driven scenarios. As illustrated in Fig.~\ref{fig:evaluate_fire}, these effects are attributed to three key factors. First, visual impairment leads to perception failures. As shown in Fig.~\ref{fig:evaluate_fire}(a), smoke occlusion causes the confidence score of a target object (Chair) to drop significantly ($0.87 \to 0.15$), leading to missed detections. Conversely, Fig.~\ref{fig:evaluate_fire}(d) demonstrates how visual noise triggers false positives, where the agent incorrectly detects a toilet (score $0.71$) in a smoke-filled corridor. Second, these perception errors destabilize exploration. While the agent in normal conditions follows a structured path (Fig.~\ref{fig:evaluate_fire}(b)), the agent under fire scenarios (Fig.~\ref{fig:evaluate_fire}(e)) exhibits erratic trajectories and fails to expand the map frontier effectively due to high sensing uncertainty. Third, the lack of hazard-aware reasoning results in unsafe navigation. Despite the presence of high-temperature zones shown in the hazard map (Fig.~\ref{fig:evaluate_fire}(c)), the standard planner fails to account for these risks. Consequently, as depicted in Fig.~\ref{fig:evaluate_fire}(f), the agent generates a path that traverses directly through hazardous areas. Our future work will include a comprehensive evaluation of \system, with particular emphasis on robustness across diverse fire scenarios.

\section{Conclusion}
In this work, we presented \system, a hazard-aware multi-agent cooperative navigation framework specifically designed for indoor fire-disaster response. Our evaluation shows that most existing cooperative navigation systems degrade substantially in fire-driven environments. Our simulation further confirms that robust perception, effective information sharing, and hazard-aware planning are crucial for reliable and efficient multi-agent SAR. By integrating multi-modal perception with VLMs, \system has the potential to significantly improve multi-agent navigation in challenging indoor fire scenarios.  
In future, we will implement and evaluate \system in both simulated and real-world fire scenarios.
\vspace{-10pt}

\bibliographystyle{IEEEtran}
\bibliography{references}

\end{document}